\documentclass[runningheads]{llncs}
\usepackage[T1]{fontenc}
\usepackage{hyperref}
\usepackage{amsfonts, amsmath, amssymb}
\usepackage{xcolor}
\usepackage{graphicx,verbatim}
\usepackage{multirow, booktabs}
\usepackage{makecell} 
\usepackage{array}
\usepackage{cleveref}
\usepackage{pifont}

\begin{document}

\title{MaLViL: Multi-axis Low-rank Vision-LSTM for Medical Image Segmentation}

\author{
    Afshin Bozorgpour\inst{1} \and
    Sina Ghorbani Kolahi\inst{2} \and
    Moein Heidari\inst{3} \and
    Ilker Hacihaliloglu \inst{3} \and
    Dorit Merhof\inst{1,4}
}
% index{Bozorgpour, Afshin}
% index{Ghorbani Kolahi, Sina}
% index{Heidari, Moein}
% index{Hacihaliloglu, Ilker}
% index{Merhof, Dorit}

\institute{
Faculty of Informatics and Data Science, University of Regensburg, Germany\\
\and % Sina
Independent Computer Science Researcher, Tehran, Iran
\and % Moein & Ilker
University of British Columbia, Vancouver, BC, Canada
\and % Merhof
Fraunhofer Institute for Digital Medicine MEVIS, Bremen, Germany
\\
% \email{afshin.bozorgpour@ur.de}
% \email{peter.schueffler@tum.de}
% \email{ejost@ukaachen.de}
\email{dorit.merhof@ur.de}
}

\authorrunning{A. Bozorgpour et al.}
\titlerunning{MaLViL: Multi-axis Low-rank Vision-LSTM}
  
\maketitle              % typeset the header of the contribution
\begin{abstract}
Vision-LSTM (ViL) enables efficient global modeling, but its cost still scales with the number of spatial tokens, so existing segmenters confine ViL to a coarse bottleneck and lose fine anatomical detail. Rasterizing 2D features into a 1D sequence further breaks adjacency across the orthogonal scan axis. We propose MaLViL, a Multi-axis Low-rank Vision-LSTM network that extends ViL across decoder resolutions. Bidirectional low-rank ViL (Bi-LRViL) reasons on a compact orthonormal subspace and preserves detail through an orthogonal residual; scale-aware SaLViL restores cross-axis neighbors before serialization; and a Cross-Directional Mixer (CDM) fuses orthogonal horizontal and vertical traversal paths. Statistics-Guided Skip Modulation (SGSM) further retains boundary cues in encoder skips. On skin-lesion, ultrasound, and multi-organ CT benchmarks, MaLViL achieves competitive or state-of-the-art segmentation accuracy, while reducing ViL operator memory by up to $83\times$ at fine decoder resolutions.
Code is available at: \href{https://github.com/xmindflow/MaLViL}{\texttt{github.com/xmindflow/malvil}}.

{\emergencystretch=1em\keywords{Medical image segmentation \and Vision-LSTM (xLSTM) \and Low-Rank Approximation \and Multi-Scale Feature Integration}}
% Authors must provide keywords and are not allowed to remove this Keyword section.
\end{abstract}
%
%
%

%%%%%%%%%%%%%%% Introduction %%%%%%%%%%%%%%%%%
\section{Introduction}\label{sec:intro}
Medical image segmentation (MIS) demands both precise delineation of fine anatomical boundaries and an understanding of long-range context~\cite{rahman2024emcad}. 
Convolutional neural networks (CNNs) provide strong local inductive biases, but their effective receptive fields can limit global structural reasoning~\cite{kolahi2024msa}. 
Vision Transformers (ViTs) address this limitation through self-attention, although its quadratic dependence on the number of image tokens is costly for high-resolution dense prediction~\cite{chen2021transunet}. 
More recently, state-space models such as Mamba have offered linear sequence modeling~\cite{gu2024mamba,ruan2024vm}; nevertheless, preserving two-dimensional locality while efficiently modeling global context remains an open challenge.

Vision-LSTM (ViL)~\cite{alkin2024vision}, which adapts the matrix-memory mechanism of xLSTM~\cite{beck2024xlstm} to visual sequences, provides an appealing alternative through bidirectional global modeling. 
Its application to dense segmentation, however, is constrained by the cost of processing long spatial sequences~\cite{chen2024xlstm}. 
For a feature map with $N{=}HW$ tokens and channel width $d$, a ViL block incurs a sequence-dependent cost of $O(Nd^2)$, making shallow, high-resolution decoder stages particularly demanding in computation and activation memory. 
Existing designs therefore tend to concentrate ViL processing at low-resolution stages, limiting its contribution to multi-scale reconstruction. 
Moreover, rasterizing a two-dimensional feature map into a one-dimensional sequence separates spatial neighbors orthogonal to the scan direction, weakening the locality required for boundary-sensitive segmentation.

To address these limitations, we propose MaLViL, a Multi-axis Low-rank Vision-LSTM architecture that extends ViL reasoning across decoder resolutions. 
Its central component, scale-aware low-rank ViL (SaLViL), projects dense spatial tokens onto a compact orthonormal subspace of rank $p\ll N$, applies bidirectional ViL to the resulting sequence, and lifts the response back to the image grid. 
Although projection and reconstruction retain an $O(Np)$ cost, the expensive ViL operation is performed on $p$ rather than $N$ tokens. 
A learnable channel-wise gate balances global subspace reasoning with information in the orthogonal residual.
SaLViL further splits channels into multi-scale groups and, before
rasterization, applies axis-aligned convolutions across the direction
orthogonal to the ViL scan so that neighbors broken by
flattening remain accessible. It is embedded in the Cross-Directional
Mixer (CDM), which runs SaLViL on two rotated views (horizontal and
vertical traversal paths) and fuses their shared response with
cross-directional disagreement to recover oriented boundaries. 
In parallel, Statistics-Guided Skip Modulation (SGSM) separates smooth and high-frequency encoder information, suppressing semantic clutter without discarding useful boundary cues. 
Together, these components couple compact global modeling with explicit spatial and frequency-aware refinement.

MaLViL is evaluated across skin-lesion, breast-ultrasound, and multi-organ CT segmentation benchmarks against representative CNN-, Transformer-, state-space-, and ViL/xLSTM-based methods. 
Direct comparison with bottleneck- and encoder-based ViL variants assesses whether low-rank modeling throughout the decoder offers benefits beyond inserting the same block family at low resolution. 
The main contributions are:
\ding{182} \textit{Multi-resolution low-rank ViL:} an orthogonal subspace formulation that performs the costly bidirectional ViL operation on $p$ compressed tokens and uses a learnable high-frequency residual gate to preserve local detail;
\ding{183} \textit{Spatial- and frequency-aware refinement:} SaLViL and CDM restore cross-directional context, while SGSM suppresses irrelevant skip responses without discarding boundary information; and
\ding{184} \textit{Comprehensive validation:} multi-modal experiments demonstrate consistent improvements over diverse segmentation architectures, supported by component ablations and operator-level memory analysis.

%%%%%%%%%%%%%%% Method %%%%%%%%%%%%%%%%%%%%%%%
\section{Methodology}\label{sec:method}
\begin{figure*}[t]
  \centering
  \includegraphics[width=\textwidth]{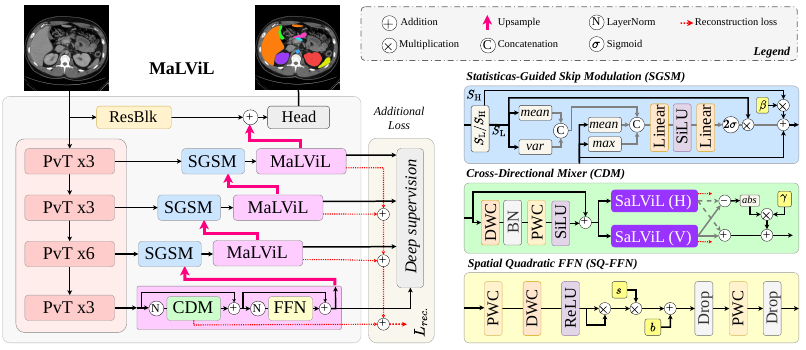}
  \caption{\textbf{MaLViL architecture.} A hierarchical encoder feeds four
  MaLViL Blocks; the three encoder skips are fused by SGSM. Each
  block contains a Cross-Directional Mixer (CDM) followed by a Spatial
  Quadratic FFN (SQ-FFN). The dashed path denotes the auxiliary low-rank
  reconstruction loss.}
  \label{fig:arch}
\end{figure*}

% =========================
% Methodology
% =========================
\label{sec:overview}
\noindent\textbf{Overview.}
MaLViL is a hierarchical encoder--decoder for medical image segmentation
(\Cref{fig:arch}). A hierarchical encoder produces features
$\{E_i\}_{i=1}^{4}$ at resolutions from $1/4$ to $1/32$ of the input.
Four MaLViL Blocks then reconstruct the prediction from coarse to fine,
and Statistics-Guided Skip Modulation (SGSM) fuses the three available
encoder skips:
\begin{equation}
 D_4=\mathcal{M}_4(E_4),\qquad
 D_i=\mathcal{M}_i\!\left(\mathrm{SGSM}
       \left(\mathcal{U}(D_{i+1}),E_i\right)\right),\ i=3,2,1,
 \label{eq:decoder}
\end{equation}
where $\mathcal{M}_i$ and $\mathcal{U}$ denote a MaLViL Block and
upsampling, respectively. A shallow residual feature of the input is
added after the final upsampling before the prediction head.

\begin{figure*}[!tbh]
  \centering
  \includegraphics[width=\textwidth]{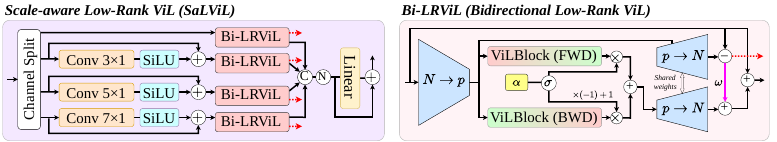}
  \caption{\textbf{SaLViL and Bi-LRViL.} Left: scale-specific
  branches recover orthogonal neighbors before Bi-LRViL sequence modeling.
  Right: rank-$p$ projection with opposite forward/backward ViL
  scans and gated fusion ($\alpha$, $\omega$).}
  \label{fig:salvil}
\end{figure*}

\noindent\textbf{MaLViL Block.}
For tokens $X\in\mathbb{R}^{B\times N\times C}$, $N=HW$, the block
alternates directional context modeling and local spatial refinement
within a pre-normalized residual formulation:
\begin{equation}
 \tilde X=X+\gamma_c\odot\mathrm{CDM}(\mathrm{LN}(X)),\qquad
 Y=\tilde X+\gamma_f\odot\mathrm{SQFFN}(\mathrm{LN}(\tilde X)),
 \label{eq:malvil_block}
\end{equation}
where $\gamma_c$ and $\gamma_f$ control the contributions of the two
residual paths. This organization first establishes long-range,
multi-directional context and then restores local spatial detail.

\noindent\textbf{Bidirectional Low-Rank ViL (Bi-LRViL).}
Applying ViL to all $N$ spatial tokens is costly at shallow decoder stages.
Bi-LRViL instead represents the spatial domain using a learned orthonormal basis
$V\in\mathbb{R}^{N\times p}$, $p\ll N$, with $V^\top V=I$.
Omitting the batch dimension, projection and lifting are
\begin{equation}
 Z=V^\top X\in\mathbb{R}^{p\times C},\qquad
 \hat X=VZ=VV^\top X,\qquad X_\perp=X-\hat X .
 \label{eq:bilrvil_proj}
\end{equation}
The expensive sequence operation therefore acts on $p$ coefficients
rather than all $N$ tokens, while $X_\perp$ preserves spatial information
outside the compact subspace. ViL is then applied only to the $p$
compressed coefficients. Along this serialized path, opposite forward
and backward traversals capture context from both ends of the sequence;
their outputs are mixed by a channel-wise gate $\alpha=\sigma(a)$:
\begin{equation}
 \bar Z=\alpha\odot\mathrm{ViL}_{f}(Z)
 +(1-\alpha)\odot\mathrm{ViL}_{b}(Z),\quad
 Y=X+V(\bar Z-Z)-\omega\odot X_\perp ,
 \label{eq:bilrvil_update}
\end{equation}
where $\omega=\sigma(w)$ down-weights the orthogonal residual
$X_\perp$ when injecting the global update, preserving fine local structure that lies outside the low-rank subspace. The reconstruction regularizer $\ell_{\mathrm{rec}}=\mathrm{MSE}(X,\hat X)$ encourages the learned basis to retain informative spatial variation.

\noindent\textbf{SaLViL.}
Rasterizing a $H{\times}W$ map into $N{=}HW$ tokens preserves adjacency
along the scan direction but separates neighbors across the orthogonal
axis. SaLViL recovers this lost two-dimensional structure before
low-rank sequence modeling (\Cref{fig:salvil}). Channels are partitioned
into scale-specific groups with complementary receptive fields
(e.g., $1{\times}1$, $3{\times}1$, $5{\times}1$, $7{\times}1$ in the
current orientation). Within each group, lightweight axis-aligned
convolutions are applied before serialization to aggregate
context across the axis opposite to the ViL traversal, thereby
bridging neighbors that would otherwise be distant in the flattened
sequence. The enriched features are then rasterized and passed to
Bi-LRViL, which performs global reasoning on only $p\ll N$ subspace
coefficients per branch. Branch outputs are concatenated and adaptively
remixed with the input.

\noindent\textbf{Cross-Directional Mixer (CDM).}
A single rasterization path is inherently anisotropic: row-major scanning
privileges one axis over the other. CDM therefore evaluates SaLViL on two
orthogonal traversal paths (the native map and a $90^\circ$-rotated
view), so horizontal and vertical neighborhoods are serialized and reasoned
over symmetrically. Denoting the two responses by $F_H$ and $F_V$, CDM
retains their shared structure while restoring oriented detail suppressed
by plain averaging:
\begin{equation}
 \mu=\frac12(F_H+F_V),\qquad
 A=\frac12\bigl(|F_H-\mu|+|F_V-\mu|\bigr),\qquad
 F_{\mathrm{CDM}}=\mu+\gamma_h\odot A .
 \label{eq:cdm}
\end{equation}
Because $A=\frac12|F_H-F_V|$ for two views, the learned scale
$\gamma_h$ re-injects cross-directional disagreement as a
direction-sensitive correction. A lightweight $3{\times}3$ residual
pathway further supplies isotropic neighborhood context.

\noindent\textbf{Statistics-Guided Skip Modulation (SGSM).}
Given decoder feature $D$ and encoder skip $S$, SGSM decomposes the skip as
$S_l=\mathrm{AvgPool}_{3\times3}(S)$ and $S_h=S-S_l$. A compact MLP
uses first- and second-order statistics of the smooth skip and decoder
context to predict a channel gate:
\begin{equation}
 q=[\mu(S_l),\mathrm{var}(S_l),\mu(D),\max(D)],\qquad
 g=2\sigma(\mathrm{MLP}(q)).
 \label{eq:sgsm_gate}
\end{equation}
Fusion is
\begin{equation}
 \mathrm{SGSM}(D,S)=D+g\odot S_l+\beta\odot S_h ,
 \label{eq:sgsm}
\end{equation}
where $g\in[0,2]$ adaptively suppresses or enhances smooth semantic
content, and the learned channel scale $\beta$ controls the propagation
of high-frequency boundary detail. This decomposition prevents strong
encoder activations from overwhelming the decoder while retaining useful
fine structure.

\noindent\textbf{Spatial Quadratic FFN (SQ-FFN) and objective.}
SQ-FFN complements global reasoning with a lightweight local transformation.
After channel expansion and depthwise spatial mixing, a learnable quadratic
activation $s\odot\mathrm{ReLU}(x)^2+b$ strengthens salient local responses
before projection to the original width. The network is optimized using
\begin{equation}
 \mathcal L=\mathcal L_{\mathrm{seg}}+\lambda\mathcal L_{\mathrm{rec}},
 \qquad
 \mathcal L_{\mathrm{rec}}=\frac14\sum_{i=1}^{4}
 \overline{\ell}_{\mathrm{rec}}^{\,i},\quad \lambda=0.01,
 \label{eq:objective}
\end{equation}
where $\mathcal L_{\mathrm{seg}}$ is the benchmark-specific segmentation
loss and $\overline{\ell}_{\mathrm{rec}}^{\,i}$ averages reconstruction
errors over the SaLViL branches and directional responses at decoder
stage $i$.

% \input{sections/method-test}

%%%%%%%%%%%%%%% Experiments %%%%%%%%%%%%%%%%%%
\section{Experiments and Results}\label{sec:experiments}
\textbf{Datasets and Experimental Setups.}\\
\noindent All experiments are implemented in PyTorch and conducted
on a single NVIDIA A5000 GPU with 24\,GB memory. MaLViL uses an ImageNet-pretrained PVTv2-B2 encoder~\cite{wang2022pvt}; the final model is trained without deep supervision. Reported methods follow their established benchmark splits, and all retrained baselines use the same data partitions as MaLViL.

\noindent\textbf{Synapse.}
The dataset contains 30 abdominal CT volumes. Following
TransUNet~\cite{chen2021transunet}, 18 volumes are used for training and 12
for testing. MaLViL is trained for 350 epochs with AdamW, batch size 8, and
an initial learning rate of $10^{-4}$. The retrained nnU-Net
baseline~\cite{isensee2021nnunet} operates at its native
$512{\times}512$ resolution. Its complexity is measured for one 2D slice,
with one multiply--accumulate reported as one FLOP to match the convention
used in the comparison tables.

\noindent\textbf{Skin-lesion and ultrasound segmentation.}
We evaluate PH$^2$, HAM10000, ISIC 2017/2018, and BUSI using their
established splits. The ISIC~2017 split contains 1,250 training, 150
validation, and 600 test images. Skin-lesion models are trained for 40
epochs at $224{\times}224$, while BUSI training uses 50 epochs at
$256{\times}256$. Both settings use AdamW, batch size 8, an initial
learning rate of $10^{-4}$, and the preprocessing and augmentation
protocol of~\cite{bozorgpour2023dermosegdiff}.

\subsection{Comparative Results}
\begin{table*}[!t]
    \centering
    \caption{Evaluation results on the Synapse dataset (\textcolor{blue}{blue} indicates the best and \textcolor{red}{red} the second best results).}
    \resizebox{\textwidth}{!}{\begin{tabular}{l|l|l|cccccccc|cc}
    \toprule
     \multirow{2}{*}{Methods}& \multirow{2}{*}{Params}& \multirow{2}{*}{FLOPs}& \multirow{2}{*}{Spl.}&  \multirow{2}{*}{RKid.}& \multirow{2}{*}{LKid.}&  \multirow{2}{*}{Gal.}&  \multirow{2}{*}{Liv.}&  \multirow{2}{*}{Sto.}& \multirow{2}{*}{Aor.}& \multirow{2}{*}{Pan.}& \multicolumn{2}{c}{Average}\\
     \cline{12-13}
     & & & & & & & & & & &DSC$\uparrow$ &HD95$\downarrow$\\

     \midrule\midrule
     R50 U-Net ~\cite{chen2021transunet}& 30.42 M& -&85.87 &78.19 &80.60 &63.66 &93.74 &74.16 &87.74  &56.90 &74.68 &36.87\\

     % DeepLabv3+~\cite{chen2018encoderdecoder}& 35.28& -& 87.43& 74.21& 82.76& 66.51& 91.23& 73.53& 88.04& 58.32& 77.63& 39.95\\

     TransUNet~\cite{chen2021transunet}& 96.07 M& 88.91 G&85.08 &77.02 &81.87 &63.16 &94.08 &75.62 &87.23  &55.86 &77.49 &31.69\\

     % LeViT-UNet-384~\cite{xu2021levitunet}& 52.17& 25.55& 88.86& 80.25& 84.61& 62.23& 93.11& 72.76& 87.33& 59.07& 78.53& 16.84\\
     
     Swin-UNet~\cite{cao2022swin}& 27.17 M& 6.16 G& 90.66& 79.61& 83.28& 66.53& 94.29& 76.60& 85.47& 56.58& 79.13& 21.55\\
     
     % MISSFormer~\cite{huang2021missformer}& 42.46& 9.89& 91.92& 82.00& 85.21& 68.65& 94.41& 80.81& 86.99& 65.67& 81.96& 18.20\\
     
     % ScaleFormer~\cite{huang2022scaleformer}& 111.6& 48.93& 89.40& 83.31& 86.36& \textbf{74.97}& \underline{95.12}& 80.14& 88.73& 64.85& 82.86& 16.81\\
     
     % HiFormer-B~\cite{heidari2023hiformer}& 25.51 M& 8.05 G& 90.99& 79.77& 85.23& 65.23& 94.61& 81.08& 86.21& 59.52& 80.39& 14.70\\
     
     % DAEFormer~\cite{azad2023daeformer}& 48.07& 27.89& 91.82& 82.39& 87.66& 71.65& 95.08& 80.77& 87.84& 63.93& 82.63& 16.39\\
     
     % PVT-CASCADE~\cite{Rahman_2023_WACV}& 35.28& 6.40& 90.10& 80.37& 82.23& 70.59& 94.08& 83.69& 83.01& 64.43& 81.06& 20.23\\

    % Laplacian-Former~\cite{azad2023laplacian}& 27.54& 6.68& 91.91& 80.52& 84.23& \underline{71.19}&\underline{94.90}& 81.14& 86.55& 64.75& 81.90& 18.86\\
     VM-UNet~\cite{ruan2024vm}& 44.27 M& 6.52 G&  89.51&  82.76& 86.16& 69.41& 94.17& 81.40& 86.40&  58.80& 81.08& 19.21\\

     % PVT-EMCAD-B0~\cite{rahman2024emcad} & 3.92 & 0.84 & 92.66&  83.96&  87.48& 66.62& 94.57& 81.22& 87.21 & 62.00& 81.97& 17.39\\

     % RWKV-UNet-T~\cite{jiang2025rwkv} & 3.30 & 3.78 & 90.36& 83.76& 86.03& 64.18& 95.09& 80.44& 89.28 &  63.50& 81.58& 17.83\\

     PVT-EMCAD-B2~\cite{rahman2024emcad} & 26.76 M & 5.60 G & 92.17& 84.10& 88.08& 68.87& 95.26& 83.92& 88.14 & 68.51& 83.63& 15.68\\
     % WaveFormer~\cite{azad2023unlocking}& 47.01& 7.75& 91.85& 80.06& 86.66& 69.67& 94.43& 80.34& 85.31& 67.08& 81.92& 18.41\\

     MSA$^{\text{2}}$Net~\cite{kolahi2024msa} & 112.77 M& 15.56 G& \textcolor{red}{92.69}& 84.24& 88.30& \textcolor{red}{74.35}& 95.59& 84.03& \textcolor{red}{89.47}& 69.30&84.75& \textcolor{red}{13.29}\\
     
     % \textbf{EW-ViT - w$\backslash$o Contrastive}& - & - & 91.36& 82.35& 86.35& 67.18& 94.62& 77.29& 87.02& 64.58& 81.34& 15.17\\

     2D D-LKA Net~\cite{Azad_2024_WACV} & 101.64 M & 19.92 G & 91.22& \textcolor{red}{84.92}& \textcolor{red}{88.38}& 73.79& 94.88& 84.94& 88.34& 67.71& 84.27& 20.04\\
    
     % \textbf{Freq-ViT- w$\backslash$o Contrastive}& - & - & -&-& -& -& -& -& -& -& -& -\\
     
     GLM-SFNet~\cite{chen2025glm}& 14.35 M & 4.64 G & 91.98& 84.35& 87.49& \textcolor{blue}{74.78}& 95.14& 85.31& 88.32 & \textcolor{red}{71.24}& \textcolor{red}{84.82}& \textcolor{blue}{11.87}\\

     UxLSTM-Bot~\cite{chen2024xlstm}& 15.02 M & 16.56 G & 91.65& 73.56& 80.21& 64.58& 94.77& 76.78& 87.14& 58.88& 78.45& 19.71\\

     nnU-Net~\cite{isensee2021nnunet}& 46.35 M & 60.02 G & 92.04& 81.48& 86.68& 65.20& \textcolor{blue}{96.31}& \textcolor{red}{85.54}& \textcolor{blue}{90.65}& \textcolor{blue}{74.80}& 84.09& 19.82\\
     \midrule
     \textbf{MaLViL}& 37.49 M & 6.56 G & \textcolor{blue}{92.74}& \textcolor{blue}{85.71}& \textcolor{blue}{89.71}& 72.98& \textcolor{red}{96.28}& \textcolor{blue}{87.64}& 88.83 & 69.91& \textcolor{blue}{85.48}& 15.23\\

     \bottomrule
     
     \end{tabular}}
    \label{tab:synapse}
\end{table*}

\begin{figure}[t]
	\centering
	\includegraphics[width=\linewidth]{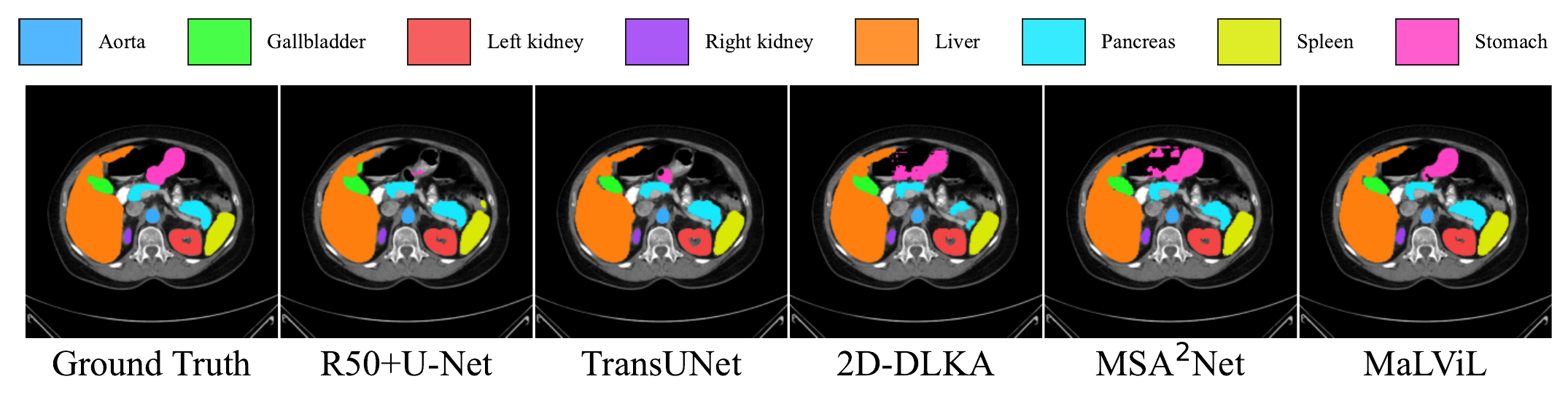}
	\caption{Qualitative comparison on the Synapse multi-organ dataset.}
	\label{fig:Comp-sin}
\end{figure}
\textbf{Synapse.}
Table~\ref{tab:synapse} reports MaLViL's performance on the Synapse
dataset, where it achieves a 0.66\% DSC improvement over the previous
best GLM-SFNet~\cite{chen2025glm}, and surpasses MSA$^2$Net~\cite{kolahi2024msa}
and 2D D-LKA Net~\cite{Azad_2024_WACV} by 0.73\% and 1.21\%, respectively,
demonstrating its effectiveness across diverse abdominal organs.
For additional reference against relevant model families, we retrain
nnU-Net~\cite{isensee2021nnunet} at $512{\times}512$ and UxLSTM-Bot~\cite{chen2024xlstm}
at $224{\times}224$, using the same split. At their respective operating
resolutions, MaLViL exceeds nnU-Net by 1.39 percentage points in mean DSC
(85.48 vs.\ 84.09); their computational profiles are 6.56 and 60.02\,G,
respectively. MaLViL also improves over the bottleneck-only UxLSTM baseline
by 7.03 percentage points (85.48 vs.\ 78.45),
supporting the benefit of low-rank ViL throughout the decoder.
Notably, nnU-Net remains highly
competitive on well-defined structures such as the aorta and pancreas,
whereas MaLViL yields more consistent gains across the smaller, harder
organs, resulting in the best overall average.
Fig.~\ref{fig:Comp-sin} further confirms MaLViL's superior multi-scale
accuracy for challenging structures such as the gallbladder, pancreas,
and stomach.

\begin{figure}[!t]
  \centering
  % \includegraphics[width=0.48\linewidth]{figs/experiments/ph2_malvil.pdf}
  % \hfill
  % \includegraphics[width=0.48\linewidth]{figs/experiments/ham_malvil.pdf}
  % \\
  \includegraphics[width=0.328\linewidth]{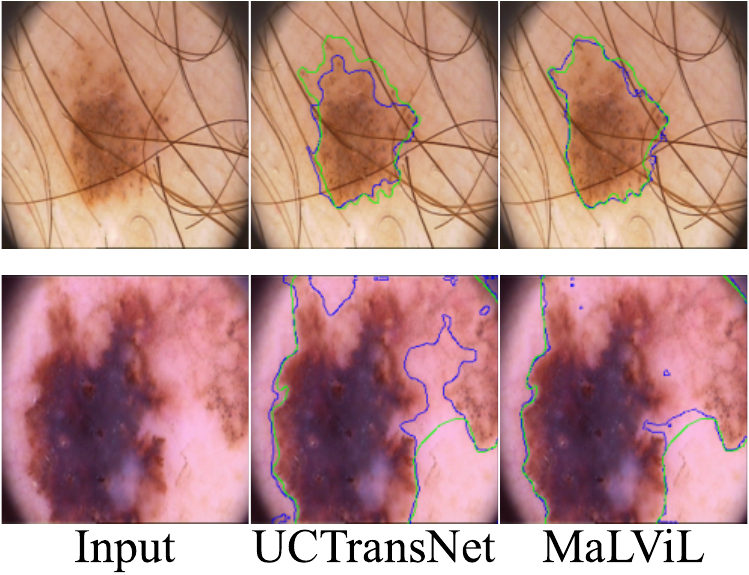}
  \hfill
  \includegraphics[width=0.328\linewidth]{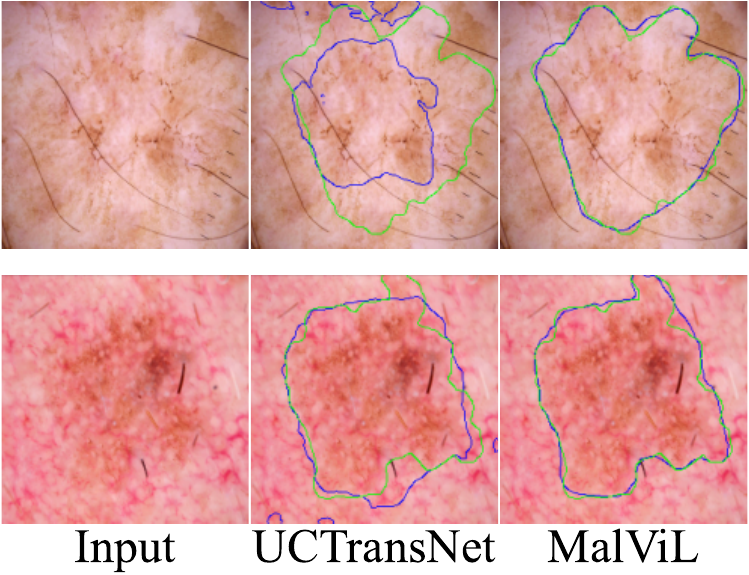}
  \hfill
  \includegraphics[width=0.328\linewidth]{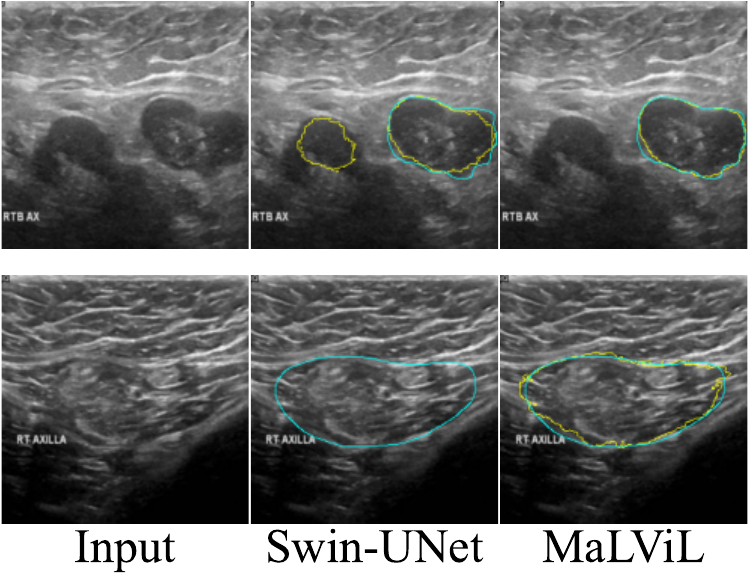}
  \caption{Qualitative segmentation on PH$^2$, HAM10000, and BUSI.}
  \label{fig:binary}
\end{figure}

\noindent\textbf{Skin Benchmarks.}
Table~\ref{tab:skin-isic-benchmarks} shows that MaLViL consistently
outperforms CNN-, Transformer-, hybrid, and Mamba-based architectures
across the evaluated skin-lesion benchmarks. It improves upon
CENet~\cite{bozorgpour2025cenet} by 0.64 and 0.46 percentage points in
DSC on PH$^2$ and HAM10000, respectively,
and attains a state-of-the-art DSC of 91.23\% on ISIC 2018. On
ISIC 2017 it reaches 92.08\% DSC, second only to
GLM-SFNet~\cite{chen2025glm} (92.18\%) while attaining the highest
accuracy (97.15\%). Fig.~\ref{fig:binary} qualitatively illustrates
the resulting lesion boundaries on PH$^2$, HAM10000, and BUSI.
\begin{table*}[!tbh]
\centering
\renewcommand{\arraystretch}{1.1}
\caption{Comparison on PH$^2$, HAM10000, ISIC'17, ISIC'18, and BUSI datasets.}
\resizebox{\textwidth}{!}{%
\begin{tabular}{|c||c||c|}

% =================== LEFT: PH2 + HAM10000 ===================
\begin{tabular}{l|cc|cc}
\toprule
\multirow{2}{*}{Methods}
  & \multicolumn{2}{c|}{PH$^2$}
  & \multicolumn{2}{c}{HAM10000} \\
\cmidrule(lr){2-3}\cmidrule(l){4-5}
  & Dice & Acc. & Dice & Acc. \\
\midrule
U-Net~\cite{ronneberger2015u}          & 89.36 & 92.33 & 91.67 & 95.67 \\
TransUNet~\cite{chen2021transunet}     & 88.40 & 92.00 & 93.53 & 96.49 \\
Swin-Unet~\cite{cao2022swin}      & 94.49 & 96.78 & 92.63 & 96.16 \\
Att-UNet~\cite{oktay2018attention}       & 90.03 & 92.76 & 92.68 & 96.10 \\
UCTransNet~\cite{wang2022uctransnet}     & 90.93 & 94.08 & 93.46 & 96.84 \\
MissFormer~\cite{huang2022missformer}     & 85.50 & 90.50 & 92.11 & 96.21 \\
CENet~\cite{bozorgpour2025cenet} & 95.04 & 97.19 & 94.71 & 97.10 \\
UxLSTM-Bot~\cite{chen2024xlstm} & 91.46 & 94.46 & 93.80 & 96.87 \\
UxLSTM-Enc~\cite{chen2024xlstm} & 91.51 & 94.42 & 93.71 & 96.89 \\
\midrule
MaLViL
               & \textbf{95.68} & \textbf{97.71}
               & \textbf{95.17} & \textbf{97.33} \\
\bottomrule
\end{tabular}
&
% =================== MIDDLE: ISIC2017 + ISIC2018 ===================
\begin{tabular}{l|cc|cc}
\toprule
\multirow{2}{*}{Methods}
  & \multicolumn{2}{c|}{ISIC2017}
  & \multicolumn{2}{c}{ISIC2018} \\
\cmidrule(lr){2-3}\cmidrule(l){4-5}
  & DSC$\uparrow$ & ACC$\uparrow$
  & DSC$\uparrow$ & ACC$\uparrow$ \\
\midrule
U-Net~\cite{ronneberger2015u}            & 89.89 & 96.13 & 88.51 & 95.39 \\
VM-UNet~\cite{ruan2024vm}          & 90.70 & 96.45 & 89.91 & 95.54 \\
VM-UNet v2~\cite{zhang2024vm}& 90.45 & 96.37 & 89.02 & 95.51 \\
DermoSegDiff~\cite{bozorgpour2023dermosegdiff}      & 91.43 & 96.72 & 89.66 & 95.75 \\
EGE-UNet~\cite{ruan2023ege}         & 90.73 & 96.42 & 88.19 & 95.10 \\
GLM-SFNet~\cite{chen2025glm} & \textbf{92.18} & 97.08 & 90.64 & 96.04 \\
UltraLight VM-UNet~\cite{wu2024ultralight} & 90.91  & 96.46  & 89.40 & 95.58 \\
UxLSTM-Bot~\cite{chen2024xlstm} & 90.00 & 96.46 & 89.92 & 96.02 \\
UxLSTM-Enc~\cite{chen2024xlstm} & 90.19 & 96.38 & 89.78 & 96.04 \\
\midrule
MaLViL
                 & 92.08 & \textbf{97.15}
                 & \textbf{91.23} & \textbf{96.75} \\
\bottomrule
\end{tabular}
&
% =================== RIGHT: BUSI (Acc removed, BUSI kept) ===================
\begin{tabular}{l|c}
\toprule
\multirow{2}{*}{Methods}
  & BUSI \\
\cmidrule(lr){2-2}
  & Dice$\uparrow$ \\
\midrule
U-Net~\cite{ronneberger2015u}         & 74.04 \\
PraNet~\cite{fan2020pranet}      & 75.41 \\
CaraNet~\cite{lou2022caranet}      & 77.34 \\
Swin-UNet~\cite{cao2022swin}      & 77.38 \\
TranFuse~\cite{zhang2021transfuse}    & 79.36 \\
% MissFormer     & 79.13 \\
U-Kan~\cite{li2025u}         & 76.40 \\
UWT-net~\cite{zhang2025uwt}        & 79.65 \\
UxLSTM-Bot~\cite{chen2024xlstm} & 72.48 \\
UxLSTM-Enc~\cite{chen2024xlstm} & 72.80 \\
\midrule
MaLViL
               & \textbf{81.76} \\
\bottomrule
\end{tabular}

\\
\end{tabular}%
}
\label{tab:skin-isic-benchmarks}
\end{table*}

\noindent\textbf{Ultrasound.}
As reported in Table~\ref{tab:skin-isic-benchmarks}, MaLViL achieves the
highest listed BUSI Dice score of 81.76\%, exceeding UWT-Net~\cite{zhang2025uwt}
by 2.11 percentage points. The examples in Fig.~\ref{fig:binary} show
accurate lesion localization and boundary delineation.

\begin{figure}[t]
\centering
\begingroup
\setlength{\tabcolsep}{1.2pt}
\begin{tabular}{@{}cccccc@{}}
\tiny Input & \tiny GT & \tiny Bi-LRViL FWD &
\tiny Bi-LRViL BWD & \tiny Bi-LRViL Out & \tiny SaLViL (H) \\
\includegraphics[width=0.158\textwidth]{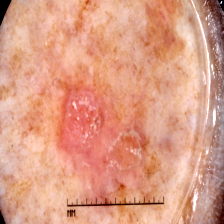} &
\includegraphics[width=0.158\textwidth]{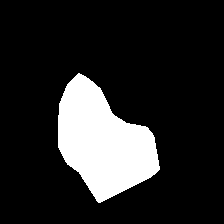} &
\includegraphics[width=0.158\textwidth]{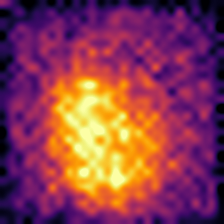} &
\includegraphics[width=0.158\textwidth]{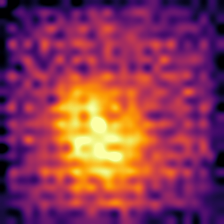} &
\includegraphics[width=0.158\textwidth]{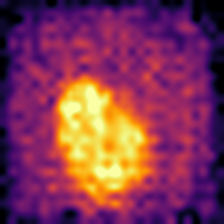} &
\includegraphics[width=0.158\textwidth]{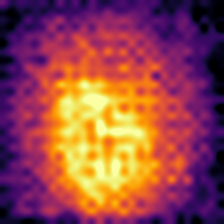} \\
\tiny SaLViL (V) & \tiny CDM Out & \tiny SGSM $S$ &
\tiny SGSM $S_{\mathrm{low}}$ & \tiny SGSM $S_{\mathrm{high}}$ &
\tiny SGSM Out \\
\includegraphics[width=0.158\textwidth]{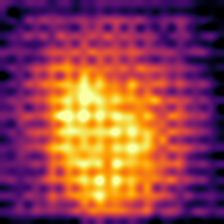} &
\includegraphics[width=0.158\textwidth]{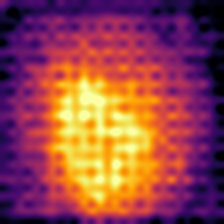} &
\includegraphics[width=0.158\textwidth]{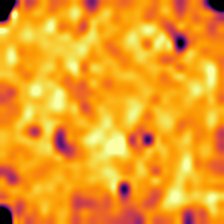} &
\includegraphics[width=0.158\textwidth]{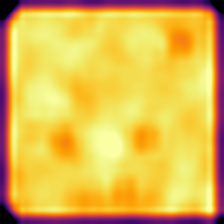} &
\includegraphics[width=0.158\textwidth]{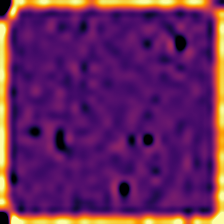} &
\includegraphics[width=0.158\textwidth]{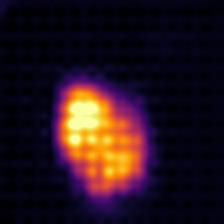}
\end{tabular}
\endgroup
\caption{\textbf{Internal features on ISIC~2017} ($28{\times}28$):
Bi-LRViL, SaLViL/CDM, and SGSM activations (normalized per map).}
\label{fig:feature-viz}
\end{figure}

\noindent\textbf{Ablation Studies.}
Figure~\ref{fig:feature-viz} visualizes a representative ISIC~2017 example
at the $28{\times}28$ decoder resolution. Bi-LRViL produces complementary
forward and backward responses, which combine into a more coherent lesion
representation. The horizontal and vertical SaLViL responses emphasize
different spatial patterns, while CDM consolidates them into a coherent target
region.
SGSM further decomposes the encoder skip $S$ into a smooth semantic component
$S_{\mathrm{low}}$ and a detail component $S_{\mathrm{high}}$ before producing
the modulated skip output. Each activation map is normalized independently
for visualization.

\begin{table}[!htbp]
\centering
\caption{\textbf{Component and efficiency analysis of MaLViL.} (a) Progressive PH$^2$ ablation at $224{\times}224$ (batch size~1); Mem. reports training-time forward/backward memory with gradients enabled, and excluded components use identity mappings. (b) Operator-level profiling of full- versus low-rank bidirectional ViL in isolation at each decoder resolution (batch size~8).}
\label{tab:ablation_malvil}
\setlength{\tabcolsep}{3pt}
\renewcommand{\arraystretch}{0.92}

\begin{minipage}[t]{0.52\textwidth}
\centering
\textbf{(a) Component contribution}\\
\resizebox{\linewidth}{!}{%
\begin{tabular}{lrrrr}
\toprule
\textbf{{\scriptsize Configuration}} &
\textbf{{\scriptsize Dice}} &
\textbf{{\scriptsize FLOPs(G)}} &
\textbf{{\scriptsize Mem.(MB)}} &
\textbf{{\scriptsize Params(M)}} \\
\midrule
Full ViL
& 93.72 & 8.83 & 1819.09 & \textbf{32.89} \\
$+$ Low-rank
& 94.30 & \textbf{5.91} & \textbf{458.11} & 33.81 \\
$+$ Bidirectional
& 94.57 & 6.05 & 482.39 & 36.21 \\
$+$ SaLViL
& 94.93 & 6.02 & 531.29 & 36.94 \\
$+$ CDM
& 95.32 & 6.60 & 630.24 & 37.34 \\
$+$ SGSM
& 95.41 & 6.60 & 635.01 & 37.49 \\
\textbf{$+$ $L_{\mathrm{rec}}$ (MaLViL)}
& \textbf{95.68} & 6.60 & 635.01 & 37.49 \\
\bottomrule
\end{tabular}}
\end{minipage}\hfill
\begin{minipage}[t]{0.46\textwidth}
\centering
\textbf{(b) Resolution efficiency}\\
\resizebox{\linewidth}{!}{%
\begin{tabular}{ccrrrrr}
\toprule
\multirow{2}{*}{\textbf{Stage}} &
\multirow{2}{*}{{\textbf{$N$}}} &
\multicolumn{2}{c}{\textbf{\rotatebox{15}{\scriptsize Mem.(MB)}}} &
\multicolumn{2}{c}{\textbf{\rotatebox{15}{\scriptsize FLOPs(G)}}} &
\multirow{2}{*}{{\textbf{Reduction}}} \\
\cmidrule(lr){3-4}\cmidrule(lr){5-6}
& & Full & LR & Full & LR & \\
\midrule
1 & $56^2$ & 10796.4 & \textbf{129.5} & 14.17 & \textbf{0.55} & $\mathbf{83\times}$ \\
2 & $28^2$ & 740.2 & \textbf{62.7} & 2.01 & \textbf{0.29} & $12\times$ \\
3 & $14^2$ & 102.8 & \textbf{48.0} & 0.67 & \textbf{0.34} & $2.1\times$ \\
4 & $7^2$ & 56.3 & \textbf{53.8} & 0.34 & \textbf{0.27} & $1.0\times$ \\
\bottomrule
\end{tabular}}
\end{minipage}
\end{table}

\noindent\begin{minipage}{\textwidth}
\textbf{Component and efficiency analysis.}
Table~\ref{tab:ablation_malvil}(a) summarizes a progressive PH$^2$ ablation at
$224{\times}224$ (batch size~1). The memory column is measured with gradients
enabled in a training-time forward/backward profile. Low-rank projection yields
the largest efficiency gain (8.83$\rightarrow$5.91\,G FLOPs,
1819$\rightarrow$458\,MB) and improves Dice from 93.72\% to 94.30\%.
Table~\ref{tab:ablation_malvil}(b) reports operator-level memory and computational
costs for full and low-rank bidirectional ViL in isolation at each decoder
resolution (batch size~8), rather than for the end-to-end network. At
$56{\times}56$, low-rank ViL reduces operator memory by $83\times$, illustrating
why MaLViL can apply ViL throughout the decoder where full-sequence modeling
becomes impractical.
\end{minipage}

%%%%%%%%%%%%%%% Conclusions %%%%%%%%%%%%%%%%%%
\section{Conclusions}\label{sec:conclusions}
We presented MaLViL, a multi-resolution decoder that makes Vision-LSTM reasoning practical beyond the bottleneck through orthogonal low-rank projection. Bi-LRViL, SaLViL, and CDM capture global context across orthogonal traversal paths, while SGSM retains informative boundary cues during skip fusion. Experiments on skin-lesion, ultrasound, and abdominal CT segmentation demonstrate competitive or state-of-the-art accuracy, with substantial ViL memory savings at fine decoder resolutions.
% A current limitation is that the projection ranks and directional configurations are fixed by stage; adapting them dynamically to image content and extending the formulation to volumetric directional modeling are promising directions for future work.

\bibliographystyle{splncs04} % Springer LNCS bibliography style
\bibliography{ref} % Entries are in the ref.bib file

\end{document}